%% file: main.tex
\documentclass[10pt,twocolumn,letterpaper]{article}

\usepackage[pagenumbers]{cvpr} 

\input{preamble}
\definecolor{cvprblue}{rgb}{0.21,0.49,0.74}
\usepackage[pagebackref,breaklinks,colorlinks,allcolors=cvprblue]{hyperref}

\newcommand{\myparagraph}[1]{
\noindent \textbf{#1}:
}

\def\paperID{****} 
\def\confName{CVPR}
\def\confYear{2026}

\title{TICON: A Slide-Level \underline{Ti}le \underline{Con}textualizer for \\ Histopathology Representation Learning}

\newcommand*\samethanks[1][\value{footnote}]{\footnotemark[#1]}

\author{\hspace{-2cm} Varun Belagali$^{1}$\thanks{Co-first authors,} , Saarthak Kapse$^{1}$\samethanks[1] , Pierre Marza$^2$\thanks{Co-second authors} , Srijan Das$^3$\samethanks[2] , Zilinghan Li$^4$,\\ \hspace{-1cm} Sofiène Boutaj$^2$, Pushpak Pati$^7$, Srikar Yellapragada$^1$, Tarak Nath Nandi$^4$, Ravi K Madduri$^{4,5}$, \\ \hspace{-1cm} Joel Saltz$^1$,  Prateek Prasanna$^1$, Stergios Christodoulidis$^2$, Maria Vakalopoulou$^{2,6}$, Dimitris Samaras$^1$  \\ \\
\hspace{-0.5cm} $^1$Stony Brook University  $^2$MICS, CentraleSupélec, Université Paris-Saclay $^3$UNC Charlotte  \\ \hspace{-0.5cm} $^4$Argonne National Laboratory  $^5$University of Chicago   $^6$Archimedes/Athena RC   $^7$Independent Researcher \\ 
\centering \small \url{https://cvlab-stonybrook.github.io/TICON/}
}

\usepackage[table]{xcolor} 
\usepackage{array}
\usepackage{multirow}
\usepackage{multicol}
\usepackage{pifont}
\definecolor{darkgreen}{RGB}{0,200,0}

\renewcommand{\baselinestretch}{1}

\begin{document}
\maketitle

\input{sec/abstract}    
\input{sec/intro}
\input{sec/related_work}
\input{sec/method}
\input{sec/experiments}

\input{sec/conclusion}

{
    \small
    \bibliographystyle{ieeenat_fullname}
    \bibliography{main}
}


\renewcommand{\baselinestretch}{1} 

\input{sec/X_suppl}

\end{document}

%% file: sec/abstract.tex
\begin{abstract}

The interpretation of small tiles in large whole slide images (WSI) often needs a larger image context. We introduce TICON, a transformer-based tile representation contextualizer that produces rich, contextualized embeddings for ``any'' application in computational pathology. Standard tile encoder-based pipelines, which extract embeddings of tiles stripped from their context, fail to model the rich slide-level information essential for both local and global tasks. Furthermore, different tile-encoders excel at different downstream tasks. Therefore, a unified model is needed to contextualize embeddings derived from ``any'' tile-level foundation model. TICON addresses this need with a single, shared encoder, pretrained using a masked modeling objective to simultaneously unify and contextualize representations from diverse tile-level pathology foundation models. Our experiments demonstrate that TICON-contextualized embeddings significantly improve performance across many different tasks, establishing new state-of-the-art results on tile-level benchmarks (i.e., HEST-Bench, THUNDER, CATCH) and slide-level benchmarks (i.e., Patho-Bench). Finally, we pretrain an aggregator on TICON to form a slide-level foundation model, using only 11K WSIs, outperforming SoTA slide-level foundation models pretrained with up to 350K WSIs.

\end{abstract}

%% file: sec/intro.tex
\section{Introduction}
\label{sec:introduction}
Histopathology whole slide images (WSIs) provide microscopic structural and cellular information about tissue samples, revealing signs of disease (\textit{e.g.,} cancer, inflammation) or normal tissue morphology. WSIs are gigapixel images, often as large as 100,000 $\times$ 100,000 pixels. 
Extracting all required information from a WSI is appealing but challenging due to computational constraints; a common strategy involves dividing it into sub-regions, or tiles. The information is then extracted for each tile separately. This traditional ``bag-of-words'' pipeline~\cite{lu2021data}, however, is suboptimal as it processes each tile in isolation, completely disregarding its surrounding neighborhood. Given that distinct tissue regions (\textit{e.g.,} tumor, stroma) often contain complementary diagnostic information, enriching each tile's representation with slide-level ``\textit{context}'' is a critical subsequent step. This process, which we refer to as \textit{tile contextualization}, is crucial for creating an enriched representation of the tiles within a WSI and motivates the work presented in this paper.

\begin{figure}[!t]
  \centering
  \includegraphics[width=1\linewidth]{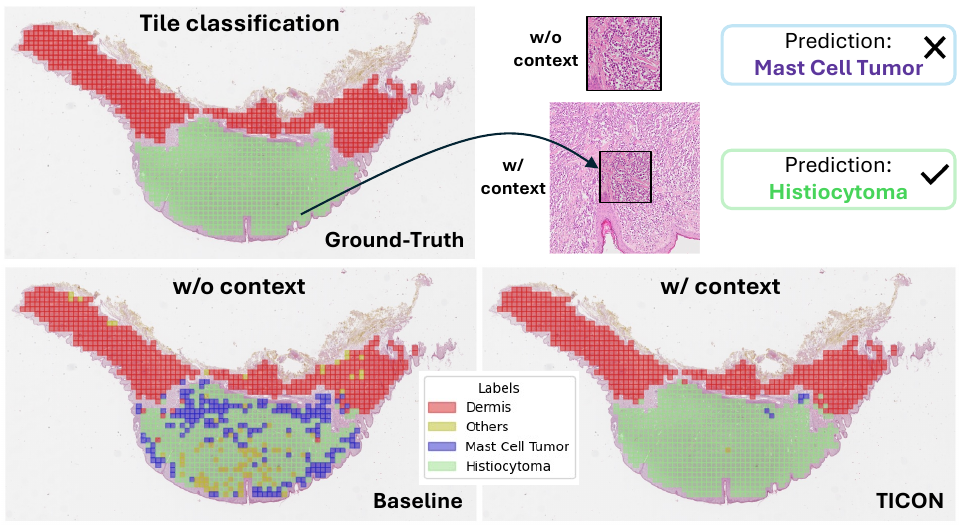}
  \caption{Tile classification in a skin cancer WSI from the CATCH dataset~\cite{catch}. Prediction without context leads to incorrect predictions (Mast Cell Tumor instead of Histiocytoma). 
  Results with our contextualizer, TICON, are much closer to the ground truth.}
  \label{fig:teaser}
  \vspace{-0.3cm}
\end{figure}

For global WSI-level tasks, the most popular approach to \textit{tile contextualization} is Multiple Instance Learning (MIL). In an MIL setting, a tile encoder, usually a foundation model trained on digital pathology data~\cite{chen2024towards, vorontsov2024foundation, zimmermann2024virchow2, hoptimus0, filiot2023scaling, filiot2024phikon, nechaev2024hibou, karasikov2025training, vaidya2025molecular, wang2024pathology, xu2024whole}, is first used to extract tile features, and then a MIL aggregator is trained on slide-level weak labels~\cite{trident-pathobench}. Here, \textit{tile contextualization} is performed only during downstream tasks, when MIL is trained with labeled WSI data. Interestingly, in the current pathology foundational model era, simpler methods such as ABMIL~\cite{ilse2018attention} that do not perform any tile contextualization,  outperform transformer-based MIL~\cite{shao2021transmil} methods that explicitly perform tile contextualization, as exhaustively benchmarked in~\cite{shaomultiple}. The lower performance of transformer-based MIL methods is potentially due to overfitting~\cite{shaomultiple}, as labeled WSIs are often limited. This suggests that the benefits of contextualizing tiles during MIL downstream tasks are limited.  For these reasons, recent works~\cite{TITAN, xu2024whole, shaikovski2024prism} have pretrained transformer-based slide-level encoders on large datasets in a self-supervised fashion. Such encoders have led to improved performance for slide-level tasks by leveraging a globally pooled representation~\cite{shaikovski2024prism,lenz2025unsupervised,TITAN}, but the explicit use of contextualized tile embeddings at the individual level has not been explored yet. This is what this paper sets out to do, and then at the second stage, studies how contextualized tile embeddings benefit the pretraining of the MIL aggregator.

Consequently, we present TICON, a transformer-based slide encoder pretrained using a masked modeling objective to learn tile contextualization. Tile embeddings contextualized by TICON improve performance on both local tile-level and global slide-level tasks. We empirically demonstrate that it is more effective to train tile-level downstream heads (linear probe/k-NN) on contextualized embeddings rather than on non-contextual ones. Similarly, MIL aggregators pretrained~\cite{jaume2024transcriptomics} on TICON's dense contextualized embeddings perform far better than their non-contextual counterparts.
TICON, when coupled with MIL aggregator pretraining (we used Tangle~\cite{jaume2024transcriptomics} in our experiments), yields a state-of-the-art slide-level foundational model, despite being trained on only 11000 WSIs.

Previous slide-level encoders~\cite{TITAN, gigapath, shaikovski2024prism} which do contextualization, aggregate information extracted from a single tile encoder. However, different tile-level encoders capture different representations with varying semantics. This is due to their different training schemes, \textit{e.g.,} vision-only DINOv2~\cite{oquab2023dinov2}-style vs vision-language CLIP~\cite{radford2021learning}-style, diverse pre-training datasets, and different neural architectures. This diversity results in a \textit{fragmented} downstream ecosystem: different encoders produce different embeddings and, consequently, excel on different downstream benchmarks. This motivates the need for a unified model that can perform dense contextualization for any tile encoder without resorting to independent contextualizers. Hence, dense contextualization of tile-level features coming from ``\textit{any}'' of those multiple tile-level encoders would conveniently tailor representations to the downstream task at hand. 

\begin{figure}[!t]
\centering
    \includegraphics[width=0.9\linewidth, clip=true, trim={0mm 4mm 0mm 0mm}]{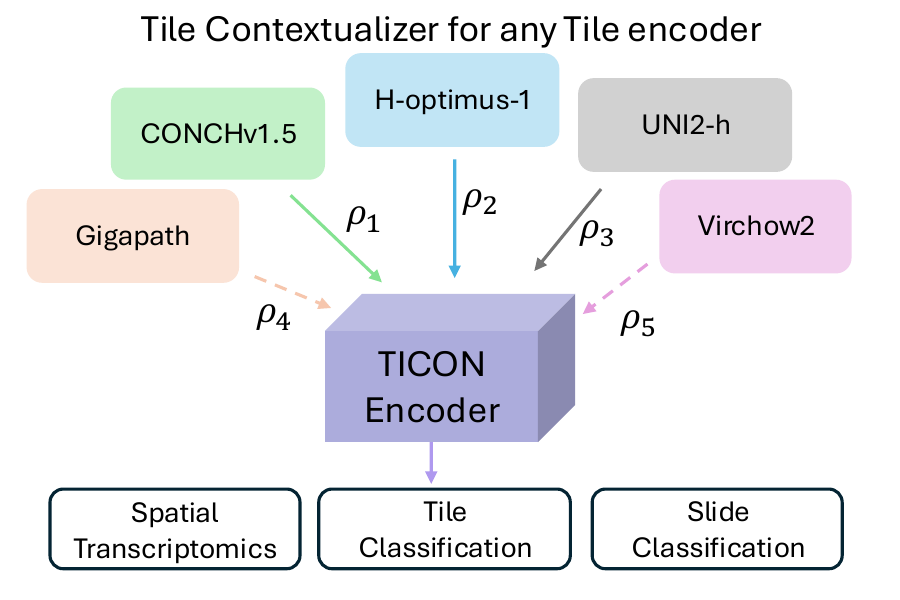} 
    \caption{TICON: An Omni Tile Contextualizer that can contextualize embeddings from any tile encoder. (---) represent input projectors for tile encoders used in pretraining. (- -) represent input projectors used in adapting TICON to new tile encoders.
    } 
    \label{fig:method_mini}
      \vspace{-0.3cm}

\end{figure}
To achieve this, we develop an ``\emph{omni}'' version of TICON using a single shared contextualization transformer that integrates multiple tile encoders.
Specifically, at each training iteration, we randomly select one tile encoder to generate the input grid of tile embeddings. A portion of these embeddings is then masked. TICON is trained to reconstruct the features for these masked tile locations. Critically, the model is pretrained to predict the target embeddings as extracted from all available tile encoders (including the embedding generated by the input tile encoder). This pretraining strategy allows TICON to effectively contextualize tile embeddings from any of the tile encoders seen during pretraining. Additionally, we show that TICON can be adapted to process embeddings from new, unknown tile-encoders by training only a light-weight input MLP projection.

An interesting property of our tile contextualizer has emerged in our experiments. Even on tile-level tasks without any context, performance improves when passing a single tile embedding through TICON. This indicates that contextualization as a pretext task helps the model to learn how to extract more relevant information from a single tile.

\noindent In summary:
\begin{itemize}
    \item We present TICON, a tile contextualizer pre-trained using masked modeling to contextualize tile embeddings from any tile encoder.

    \item We demonstrate the effectiveness of TICON across a diverse set of tile-level and slide-level tasks, showing the benefit of contextualizing tile embeddings. In fact, TICON can even operate on new tile encoders not seen during its pre-training.

    \item We show that a foundation model built with TICON outperforms other state-of-the-art slide-level foundation models.

\end{itemize}

%% file: sec/related_work.tex
\section{Related Work}

\textbf{Tile-level encoders} are pre-trained on large histopathology datasets in a self-supervised fashion. We can divide them into two main categories: (i) vision-only models trained with an iBOT~\cite{zhou2021ibot} or DINOv2~\cite{oquab2023dinov2}-style objective~\cite{chen2024towards, vorontsov2024foundation, zimmermann2024virchow2, hoptimus0, filiot2023scaling, filiot2024phikon, nechaev2024hibou, karasikov2025training, vaidya2025molecular, wang2024pathology, xu2024whole} and (ii) vision-language models trained with a CLIP-\cite{radford2021learning}-style pre-training~\cite{lu2024visual, TITAN, zhou2024knowledge, ikezogwo2023quilt, huang2023visual, xiang2025vision, shaikovski2024prism}. Among both groups, models differ in terms of their sizes and the characteristics of their pre-training datasets. A large body of work has studied their differences and compared their performances~\cite{wolflein2023benchmarking, kang2023benchmarking, neidlinger2024benchmarking, gustafsson2024evaluating, alfasly2024foundation, gatopoulos2024eva, jaume2024hest, breen2025comprehensive, lee2025benchmarking, majzoub2025good, alfasly2025validation, zhang2025accelerating, campanella2025clinical, ma2025pathbench, marza2025thunder, gindra2025large}, showing that their strengths and weaknesses differed based on the downstream tasks and data distributions. Such tile-level encoders are an essential building block of current slide-level models as a way to extract relevant information from sub-regions in high-resolution WSIs.

As tile-level encoders extract different information, we propose to train a unique slide-level encoder that can contextualize tile embeddings coming from different encoders, providing more flexibility at downstream time.

\textbf{Downstream aggregators} are a popular way to fuse tile-level embeddings coming from tile-level foundation models. As WSIs often come only with slide-level labels, the MIL framework, where a slide is treated as a bag of tile instances, has been extensively used. Methods such as ABMIL~\cite{ilse2018attention} compute attention scores for each tile independently from its embedding. More recent approaches with transformer-based MILs (such as TransMIL~\cite{shao2021transmil}), integrate explicit contextualization through a self-attention mechanism~\cite{vaswani2017attention}. However, the latter is learned for downstream data, using labeled WSIs, which are often limited in number. 
Indeed, recent work~\cite{shaomultiple} has exhaustively benchmarked that ABMIL outperforms many other sophisticated MIL methods, on downstream slide-level tasks, suggesting a limitation of downstream contextualization learning. 

\textbf{Slide-level encoders} were introduced to learn to aggregate tile embeddings from larger unlabeled datasets~\cite{TITAN, vaidya2025molecular, wang2024pathology, xu2024whole, shaikovski2024prism, lenz2025unsupervised, kapse2025gecko}, overcoming the limitations of supervised aggregators. Slide-level models~\cite{TITAN, vaidya2025molecular, wang2024pathology, xu2024whole, shaikovski2024prism} are pre-trained with iBOT, CLIP, or Masked AutoEncoding (MAE)~\cite{he2022masked} objectives. Among them, TITAN~\cite{TITAN} is a multi-modal slide-level encoder trained with an iBOT and vision-language alignment objective, whereas Gigapath~\cite{xu2024whole} is trained with a masked autoencoding (MAE) objective. CrossMAE~\cite{fu2024rethinking}, a recent model in natural imaging, utilizes a cross-attention-based decoder that provides flexibility for partial prediction during MAE pretraining. Motivated by this, we adopted a cross-decoder for our pretraining because it is well-suited for unstructured tissue regions in histopathology slides.

Previous work on slide-level encoders showed increased performance compared with downstream aggregators, however the benefits of their implicitly conducted tile contextualization have not been studied for tile-level tasks. We take one step further in this paper by validating the benefits of tile contextualization, both at the local tile level and the global slide level. 

\textbf{Multi-teacher distillation} pertains to distilling the knowledge of multiple teacher models into a student encoder~\cite{ranzinger2024radio, sariyildiz2024unic, sariyildiz2025dune}. GPFM~\cite{ma2024towards} has explored this direction in histopathology. More specifically, GPFM combined a self-supervised DINOv2 objective with a multi-teacher distillation loss to train a tile-level student encoder. Multi-teacher distillation approach has not been extensively studied for slide-level encoders. COBRA~\cite{lenz2025unsupervised} incorporates multiple tile encoders as inputs in contrastive pretraining to learn encoder-agnostic attention pooling. However, during inference, it uses only the learned attention to pool the original, non-contextual tile embeddings.

In this paper, we extend the concept of multi-teacher distillation to train a unified slide-level contextualizer to reconstruct tile-level representations from different tile-level foundation models.

%% file: sec/method.tex
\label{sec:related_work}

\section{Preliminaries}

\myparagraph{Tile Embedding Extraction} A Whole Slide Image (WSI) is first partitioned into an $M \times N$ grid of non-overlapping tiles ($X$), each of size $t \times t$ pixels. Each tile is then processed $\textit{independently}$ by an off-the-shelf pre-trained tile encoder ($\phi_i$) to yield a $d_i$-dimensional embedding. This process results in a feature grid, $\phi_i(X) \in \mathbb{R}^{M \times N \times d_i}$, composed of $\textit{non-contextualized}$ tile embeddings, where each embedding lacks information about its surrounding tiles. In this work, we utilize $\textit{multiple}$ such tile encoders, $\{\phi_i\}_{i=1}^T$, to extract distinct, parallel representations for each WSI.

\myparagraph{Histopathology Downstream Tasks} Following embedding extraction, these non-contextualized features can be used in various downstream tasks, such as spot-level gene expression prediction~\cite{jaume2024hest}, tile-level classification~\cite{marza2025thunder,gatopoulos2024eva}, and WSI-level classification~\cite{trident-pathobench,ma2025pathbench}. Traditionally, a Multiple Instance Learning (MIL) aggregator~\cite{ilse2018attention,lu2021data} is trained for WSI-level tasks, while a simple projection head is used for spot- and tile-level predictions. Consequently, spot- and tile-level predictions are typically made without considering neighborhood context. Furthermore, while sophisticated MIL aggregators can be transformer-based~\cite{shao2021transmil,reisenbuchler2022local}, the limited availability of labeled data often results in simpler aggregators, such as Attention-based MIL (ABMIL~\cite{ilse2018attention}), yielding more robust performance. This reliance on non-contextual tile processing and data-limited aggregators highlights a central gap: it remains an open question whether ``bag-of-words" approaches are sufficient, or if effective tile contextualization is necessary to model complex histopathological patterns. In this paper, we provide evidence for the latter, showing that tile contextualization is useful and thus suggesting its low benefit in downstream MIL settings might be due to a lack of data diversity.

\begin{center}
 \begin{figure*}[t]
 \centering
     \includegraphics[width=0.9\linewidth]{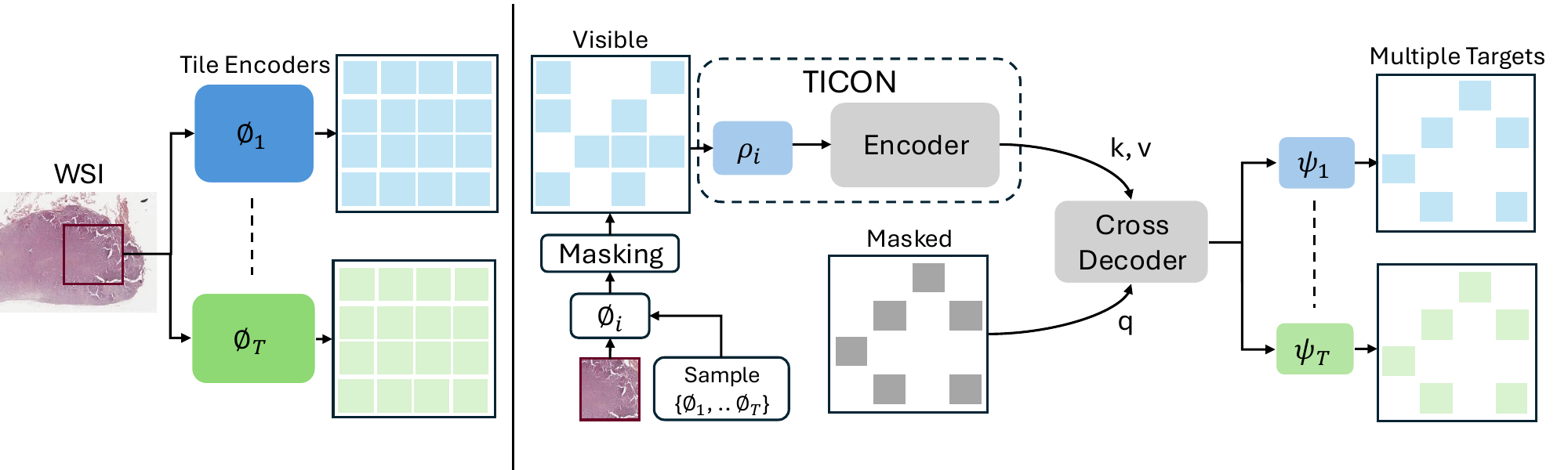} 
     \caption{Overview of the pretraining framework. \textbf{(Left)} Grid sampling and tile embedding extraction using a set of tile encoders ($\phi_1, \phi_2, \dots, \phi_T$). \textbf{(Right)} An input tile encoder ($\phi_i$) is sampled randomly at each iteration, and its embeddings are masked. The remaining visible embeddings are passed through a $\phi_i$-specific input projector ($\rho_i$) and then a shared encoder. A shared decoder, paired with output projectors specific to each tile encoder, then reconstructs the masked embeddings corresponding to all tile encoders ($\phi_1, \dots, \phi_T$).}
     \label{fig:method}
       \vspace{-0.3cm}

 \end{figure*}
 \end{center}

\section{Proposed Method}
\label{sec:method}

In this section, we introduce TICON, illustrated in Fig.~\ref{fig:method_mini}. TICON is designed to contextualize tile embeddings from any tile encoder within a WSI. We first detail the TICON architecture in Sec.~\ref{sec:ticon_architecture}. Next, we elaborate on our omni-feature masked modeling-based pretraining paradigm in Sec.~\ref{sec:ticon_pretraining}. Finally, we describe the process for adapting TICON to unseen tile encoders in Sec.~\ref{sec:ticon_adaptation}.

\subsection{TICON architecture}
\label{sec:ticon_architecture}

Our framework consists of two primary components (Fig.~\ref{fig:method_mini}): (i) input projection layers, specific to each tile encoder, and (ii) a shared, transformer-based slide-encoder. We formalize both components in detail below. 

\myparagraph{Separate Input Projectors}
For each distinct tile encoder $\phi_i$, we employ a separate input projection MLP, $\rho_i$. This projector maps the embeddings from the native dimension ($d_i$) of $\phi_i$ to a shared embedding dimension $D$, which serves as the input dimensionality for the Slide Encoder.

\myparagraph{Shared Slide Encoder}
We employ a single, transformer-based Slide Encoder, $\mathcal{E}$,  with $l$ layers, which is shared across all tile encoders. It takes as input the sequence of projected, non-contextual tile embeddings $\rho_i(\phi_i(X))$, corresponding to a WSI's tile grid $X$  processed by a specific tile encoder-input projector pair ($\phi_i$, $\rho_i$). By modeling long-range dependencies between these embeddings, $\mathcal{E}$ outputs enriched, contextualized tile embeddings as follows: 

\begin{align}
    {E}_{ctx} &= \mathcal{E}(\rho_i(\phi_i(X)))
    \label{eq:inference}
\end{align}

The versatile shared encoder $\mathcal{E}$ can process inputs originating from \textit{any} tile encoder $\phi_i$. Thus, prior knowledge of the most suitable encoder for a downstream task allows us to use TICON with that specific encoder, further enhancing its performance.

\subsection{Omni-Feature Masked Modeling}
\label{sec:ticon_pretraining}
Because TICON's target output admits no direct supervisory signal, we pretrain it in a self-supervised manner. To this end, we introduce \emph{Omni-Feature Masked Modeling (OFMM)}, a pretraining paradigm that leverages the complementary capabilities of heterogeneous tile encoders (\textit{i.e.,} omni-type features) and aligns them in a shared semantic space. By incorporating slide-level context, OFMM enhances the embeddings produced by each individual tile encoder, while promoting consistency across tile encoders. Fig.~\ref{fig:method} illustrates the pretraining framework.
OFMM operates by masking a portion of the input tile features, which are then passed through TICON. The resulting representations are fed into a cross-decoder that reconstructs multiple targets, specifically, the masked features themselves as well as the corresponding features of the same location obtained from a different tile encoder. We next detail each step of OFMM.

At each iteration, we randomly select an input tile encoder $\phi_i$ and obtain the non-contextual tile embeddings $\phi_i(X)$ for a given WSI tile grid $X$. We apply a random mask to these embeddings with a masking ratio of $m_r$. The masked tile embeddings $\phi_i^m(X)$ are discarded, while the remaining visible tile embeddings $\phi_i^v(X)$ are passed through the input projection MLP $\rho_i$ and then into the slide encoder $\mathcal{E}$. We use ALiBi~\cite{alibi} to incorporate relative 2D position encoding within the encoder's attention layers. Finally, the encoder outputs ${E}_{ctx}$, a sequence of contextualized representations for the visible tiles as follows:

\begin{align}
    \phi_i^v(X), \phi_i^m(X) &= \text{mask}(\phi_i(X), m_r) \\
    {E}_{ctx} &= \mathcal{E}(\rho_i(\phi_i^v(X)))
    \label{eq:encoder}
\end{align}

The \textbf{Cross-Decoder} $\mathcal{D}$ reconstructs the masked tile embeddings for \emph{all} target encoders $\{\phi_j\}_{j=1}^{T}$, where $T$ denotes the number of tile encoders used during pretraining. Let $m$ be the set of masked indices. We first subsample a set of prediction locations $p \subseteq m$ according to a prediction ratio $p_r$: 
\begin{equation}
    p = \operatorname{Sample}(m, p_r).
\end{equation}
The decoder input consists of mask tokens $z \in \mathbb{R}^{|p|\times d}$ formed by repeating a single learnable token $\mu \in \mathbb{R}^{d}$ once per index in $p$:
\begin{equation}
    z = \operatorname{Repeat}(\mu, p).
\end{equation}
$\mathcal{D}$ is a Transformer that cross-attends to the contextualized visible embeddings ${E}_{\text{ctx}}$ produced by the slide encoder to gather slide-level context. As in the encoder, we incorporate ALiBi positional biases within the decoder’s attention layers. The decoder outputs an intermediate representation $Z$ at the prediction locations $p$:
\begin{equation}
    Z = \mathcal{D}(z, E_{\text{ctx}}).
\end{equation}

\myparagraph{Output Projectors}
The shared representation $Z \in \mathbb{R}^{|p|\times D}$ is projected from the decoder’s embedding dimension $D$ into the feature space of each tile encoder. We employ a set of encoder-specific output heads (MLPs) $\{\psi_j\}_{j=1}^{T}$ that produce the final predicted embeddings $\mathbf{y}_j$ for each target encoder:
\begin{equation}
    \mathbf{y}_j = \psi_j(Z), \quad \forall\, j \in \{1,\ldots,T\}.
    \label{eq:output_projection}
\end{equation}

\myparagraph{Multi-Target Loss}
At each training iteration, our method uses the set of embeddings from a single sampled tile encoder, $\phi_i$, as the input. The model is then tasked with predicting the tile embeddings as generated by all available tile encoders ($\phi_1, \ldots, \phi_T$) at the prediction locations $p$.

While all encoders ($\phi_1, \ldots, \phi_T$) describe the same tile, they provide different \textquoteleft views\textquoteright{} of that tile. By training the model to predict these multiple, diverse target representations, our \textit{multi-target loss} $\mathcal{L}$ provides additional supervision. This encourages the model to learn a more robust contextual understanding, as it must bridge the gap between these varied yet semantically equivalent representations. We train all model parameters ($\{\rho_i\}_{i=1}^T, \mathcal{E}, \mathcal{D}, \{\psi_j\}_{j=1}^T$) end-to-end by minimizing $\mathcal{L}$, defined as:

\begin{align}
    \mathcal{L} &= \sum_{j=1}^{T} \text{1-cos}(\mathbf{y}_j, \phi_j^p(X))
    \label{eq:loss}
\end{align}
where $\mathbf{y}_j$ is the predicted embedding for encoder $\phi_j$ (from Eq.~\ref{eq:output_projection}), and $\phi_j^p(X)$ represents the ground truth embeddings from encoder $\phi_j$ at the prediction locations $p$.

\myparagraph{Training Strategy}
Note that, given the large number of tiles per WSI, we adopt a train-short-test-long paradigm~\cite{press2021train, TITAN}. TICON is pretrained on smaller, square grids of $K \times K$ tiles (where $K < M, N$). Leveraging ALiBi~\cite{press2021train} position encoding, our model effectively extrapolates to the full WSI resolution at inference time.

\subsection{Adapting to Unseen Tile Encoders}
\label{sec:ticon_adaptation}

To allow TICON to operate on embeddings from a novel, unseen tile encoder ($\phi_u$), we introduce a simple and parameter-efficient adaptation strategy. Our approach involves freezing the core shared components: the slide encoder ($\mathcal{E}$) and the cross-decoder ($\mathcal{D}$). We then introduce and train two new, lightweight components: an input projector ($\rho_u$) and an output projector ($\psi_u$). These projectors are optimized using a self-reconstruction objective, analogous to the pretraining phase but simplified to a single-target loss, where the model is tasked with reconstructing the masked embeddings of $\phi_u$ itself. At inference time, TICON uses the newly trained input projector $\rho_u$ with the frozen pretrained Slide Encoder $\mathcal{E}$ to contextualize tile embeddings from $\phi_u$.

%% file: sec/experiments.tex
\section{Experiments and Results}

We first provide the implementation details for our pretraining and downstream evaluation in Sec.~\ref{sec:implementation}. We then demonstrate the effectiveness of TICON on tile-level tasks in Sec.~\ref{sec:results-tile-level}. Next, in Sec.~\ref{sec:results-slide-level}, we demonstrate the effectiveness of contextualized tile embeddings for building a SoTA slide-level foundation model. We also show how to adapt TICON to unseen tile encoders in Sec.~\ref{sec:results-unseen-tile-encoders}, and present ablation studies in Sec.~\ref{sec:ablations}. Please refer to appendix for additional implementation details and results.

\label{sec:experiments}
\subsection{Setup and Implementation}
\label{sec:implementation}

\myparagraph{Pretraining Data}
We utilize 11,000 whole-slide images (WSIs) from The Cancer Genome Atlas (TCGA)~\cite{tcga} for our pretraining. Following Trident~\cite{trident-pathobench} for tissue segmentation, we extract non-overlapping 512 $\times$ 512 pixel tiles ($t=512$) at 20X magnification. From these, we construct $16 \times 16$ grids ($K=16$) as ``pretraining candidates." We sample a maximum of 20 candidates per WSI, ensuring each contains at least 55\% tissue region, resulting in a total of 195K candidates. This supports our train-short-test-long paradigm: the model is pretrained on these $16 \times 16$ candidates but processes all tile embeddings from the entire WSI at inference.

\myparagraph{Tile Encoders}
For TICON, we employ three state-of-the-art tile encoders during pretraining: H-optimus-1~\cite{hoptimus1} (1536-dim), UNI2-h~\cite{uni2} (1536-dim), and Conchv1.5~\cite{conchv15} (768-dim). While Conchv1.5 natively processes 512 $\times$ 512 tiles, H-optimus-1 and UNI2-h operate on 224 $\times$ 224 inputs. To harmonize these, we generate a single 512 $\times$ 512 tile representation for H-optimus-1 and UNI-2h by average-pooling the embeddings of four constituent 224 $\times$ 224 sub-regions. This ensures all encoders produce a consistent $16 \times 16$ grid of embeddings for any given candidate. Once pretrained, TICON supports inference with H-optimus-1, UNI2-h, and Conchv1.5 tile encoders.

\myparagraph{TICON Configurations}
TICON slide encoder ($\mathcal{E}$) is composed of 6 ViT blocks, and the cross-decoder ($\mathcal{D}$) is composed of 1 ViT block, both operating with a shared embedding dimension of $D=1536$. The tile-encoder-specific input projectors ($\rho_i$) and output projectors ($\psi_j$) are implemented as 2-layer MLPs. For pretraining, we use a masking ratio $m_r=0.75$ and a prediction ratio $p_r=0.25$. We train the model for 100K iterations with a batch size of 1024, using an AdamW optimizer and a learning rate of $2 \times 10^{-4}$.

\myparagraph{Downstream Evaluation}
We evaluate TICON on a comprehensive set of 53 downstream tasks spanning both tile-level and slide-level. For tile-level tasks, we evaluate on HEST-Bench~\cite{jaume2024hest} (9 tasks) for gene expression prediction, the CATCH dataset~\cite{catch} (1 task), and the Thunder Benchmark~\cite{marza2025thunder, nechaev2025spider} (16 tasks for tile classification, combining 12 original and 4 SPIDER tasks). For slide-level evaluation, we use 27 tasks from Patho-Bench~\cite{trident-pathobench} (2 from BRACS~\cite{nechaev2025spider} and 25 from CPTAC~\cite{edwards2015cptac}). In all experiments, we strictly adhere to the evaluation protocols and reporting metrics established by each respective benchmark. We adapt the CATCH, which originally provides segmentation contours, for tile-level classification. Specifically, a tile is assigned a class label only if it is \textit{fully contained} within the corresponding contour. To ensure an unambiguous multi-class classification task, tiles that overlap with multiple contours are excluded. For more details regarding this CATCH curation, please refer to the appendix.

\subsection{Tile-level task evaluation}
\label{sec:results-tile-level}
To analyze the impact of contextualization, we test two distinct settings: (i) \textit{with slide context}, where the entire WSI is processed to generate context-enriched tile embeddings, and (ii) \textit{without slide context}, where each tile embedding is passed \textit{in-isolation} through the TICON   (denoted as TICON$_{iso}$). In this latter setting, the self-attention mechanism operates on a sequence of length one, effectively boiling down the Transformer to a deep MLP.

\subsubsection{WSI datasets}
\label{sec:wsi_datasets}

\myparagraph{Evaluation}
For HEST-Bench (gene expression prediction), we follow the established protocol of using a ridge regression model with PCA and report Pearson correlation. For CATCH (12-class skin classification), we follow the k-NN protocol and report the F1-score.

\begin{table}[!htbp]
\centering
{
\caption{Tile-level tasks with slide context. We report the Pearson correlation coefficient (PCC) for HEST-Bench and the F1-score for CATCH. For each tile encoder (e.g., H-optimus-1), we compare its baseline performance against our TICON variants: with slide-level context and without slide-level context (TICON$_{iso}$), when inferred with the corresponding tile encoder.}
\label{tab:patch_tasks_with_context}
\resizebox{0.8\columnwidth}{!}{%
\begin{tabular}{l c c}
\toprule
    \textbf{Method} & \textbf{HEST-Bench} (9 tasks) & \textbf{CATCH} (1 task) \\
    \midrule
    H-optimus-1     & 0.422 & 86.2 \\
    w/ TICON$_{iso}$ & 0.423 & 86.6 \\
   w/  TICON & \textbf{0.427} & \textbf{87.6} \\
    \midrule
    UNI2-h     & 0.414 & 85.5 \\
  w/   TICON$_{iso}$ & 0.418 & 85.8 \\
 w/    TICON  & \textbf{0.419} & \textbf{86.9} \\
    \midrule
    CONCHv1.5  & 0.379 & 81.6 \\
  w/   TICON$_{iso}$ & 0.378 & 83.2 \\
   w/  TICON  & \textbf{0.381} & \textbf{84.9} \\
\bottomrule
\end{tabular}}}
\end{table}
\myparagraph{Results}
As shown in Table~\ref{tab:patch_tasks_with_context}, we compare the baseline non-contextual tile encoders (CONCHv1.5, H-optimus-1, UNI2-h) against two TICON variants: (i) with slide-level context (TICON) and (ii) without slide context - in an isolated setting (TICON$_{iso}$). Specifically, TICON processes all tile embeddings from the WSI to model inter-tile context, while TICON$_{iso}$ processes each tile independently.
We observe that TICON consistently outperforms the non-contextual baselines (e.g., up to 0.005 on HEST and 3.3\% on CATCH), highlighting the benefits of slide-level contextualization. Furthermore, the performance gap between TICON and TICON$_{iso}$ indicates that the observed improvement is directly attributable to modeling inter-tile context. Interestingly, TICON$_{iso}$ still outperforms the original tile embeddings in most settings, suggesting that TICON’s omni-feature pretraining produces a more effective representation, even when explicit contextual information is removed.

Moreover, TICON establishes a new state-of-the-art in HEST-Bench, pushing the H-optimus-1 PCC performance from 0.422 to 0.427.
For comparison, the 0.007 performance increase observed between H-optimus-0 and H-optimus-1 (0.415 $\rightarrow$ 0.422) was obtained only after doubling the pretraining data from 500K to 1M WSIs, highlighting the magnitude of the improvement achieved here. TICON achieves a comparable 0.005 improvement by pretraining on only 11K TCGA WSIs. Similarly, in the CATCH dataset, TICON increases the baseline H-optimus-1 from 86.2\% to 87.6\%.

\subsubsection{Tile datasets}

\myparagraph{Evaluation}
From Sec~\ref{sec:wsi_datasets}, we observe that TICON provides a performance uplift even in the \textit{without slide context} setting. Motivated by this finding, we conduct a more exhaustive evaluation on the 16 tile classification tasks in the Thunder benchmark~\cite{marza2025thunder, nechaev2025spider}, for which slide-level data is not available. We also follow the k-NN protocol and report F1-scores.
\vspace{-23pt}

\begin{center}
 \begin{figure*}[t]
 \centering
     \includegraphics[width=0.9\linewidth]{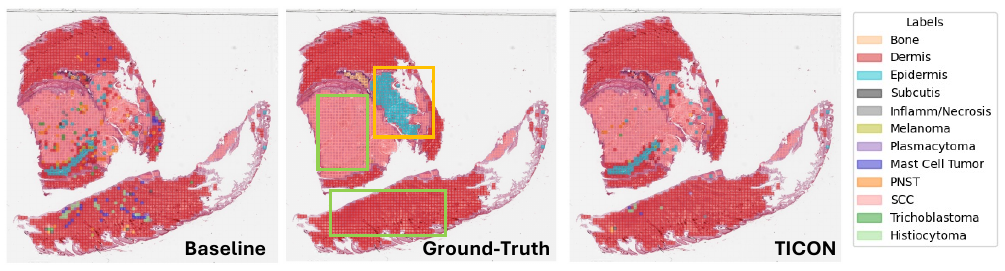} 
\caption{Visualization of tile classification results on a WSI from the CATCH dataset. The left panel (baseline) shows classification using non-contextual tile embeddings, whereas the right panel (TICON) displays classification with contextualized embeddings. TICON produces less noisy predictions and corrects many local misclassifications (\textcolor{darkgreen}{green} boxes). However, we also observe shared failure modes (\textcolor{orange}{orange} box) where both methods misclassify a region. This suggests limitations in the underlying tile encoder's features. Indeed, the latter can miss some necessary information that even contextualization cannot retrieve.}
     
     \label{fig:qualitative}
       \vspace{-0.3cm}

 \end{figure*}
 \end{center}

\myparagraph{Results}
As shown in Table~\ref{tab:patch_tasks_no_context}, TICON$_{iso}$ yields notable improvements: up to 1.4\% on the SPIDER subset (4 tasks) and 1.8\% on the Thunder (original) subset (12 tasks). These result highlights an interesting property: TICON's masked pretraining learns a superior tile-level representation, which proves beneficial even when explicit context is not available. Critically, by applying TICON$_{iso}$, we boost the performance of UNI2-h, achieving new state-of-the-art results on the Thunder (original) benchmark.

\begin{table}[!htbp]
\centering
{
\caption{Tile-level tasks without slide context. F1-score is reported for both tasks. }
\label{tab:patch_tasks_no_context}
\resizebox{0.9\columnwidth}{!}{%
\begin{tabular}{l|cc}
\toprule
    \textbf{Method} & \textbf{THUNDER-SPIDER} & \textbf{THUNDER-ORIGINAL}  \\
    & 4 tasks & 12 tasks \\ 
    
   \midrule
    H-optimus-1  & 88.4 & 80.5 \\
   w/ TICON$_{iso}$    & \textbf{89.0} & \textbf{80.8} \\
    \midrule
    UNI2-h  & 88.2 & 81.7 \\
   w/  TICON$_{iso}$    & \textbf{88.5} & \textbf{82.1} \\
    \midrule
    CONCHv1.5 & 83.7 & 78.7 \\
    w/ TICON$_{iso}$    & \textbf{85.1} & \textbf{80.5} \\
      
\bottomrule
\end{tabular}}}
\end{table}

\subsection{Building Slide-level foundation models}
\label{sec:results-slide-level}
In this section, we build a slide-level foundation model by pretraining an MIL aggregator using a Tangle-based objective~\cite{jaume2024transcriptomics} on top of TICON's contextualized embeddings. The output of this pretrained MIL aggregator serves as a global, slide-level representation for downstream applications. We train a separate MIL aggregator for each input tile encoder. The Tangle aggregator itself is an Attention-Based MIL (ABMIL~\cite{ilse2018attention}) model trained using contrastive learning between the aggregated tile embeddings and the WSI's bulk gene expression. Tangle pretraining is performed on the TCGA 10K WSI-gene pairs. We term this combined model TICON$_{{tangle}}$.

\begin{table}[t]

\centering
\caption{Evaluation of slide-level foundation models with Patho-bench~\cite{trident-pathobench}. Following Pathobench, we report balanced accuracy for BRACS and AUC for CPTAC. TICON$_{{tangle}}$ improves over Tangle across all tile encoders and outperforms other slide-level foundation models.}
\label{tab:slide-foundation-model}
\resizebox{0.8\columnwidth}{!}{
\begin{tabular}{@{}lc|cc@{}}
\toprule
\multirow{1}{*}{\textbf{Model}} &  \multirow{1}{*}{\textbf{Tile Encoder}} & 
\multicolumn{1}{c}{\textbf{BRACS}} &
\multicolumn{1}{c}{\textbf{CPTAC}} \\

& & 2 tasks & 25 tasks \\

\midrule 
TITAN~\cite{TITAN}  & CONCHv1.5~\cite{conchv15} & 60.1 & 69.4  \\
CHIEF~\cite{ding2411multimodal}  & CTransPath~\cite{wang2022transformer} & 57.9 & 66.5  \\
PRISM~\cite{shaikovski2024prism}  & Virchow~\cite{vorontsov2023virchow} & 59.0 & 66.5  \\
Madeleine~\cite{jaume2024multistain}  & CONCHV1~\cite{lu2024visual} & 57.2 & 69.4  \\
Feather~\cite{tang2024feature}  & CONCHv1.5~\cite{conchv15}  & 56.8 & 64.7  \\

COBRA~\cite{lenz2025unsupervised} & Virchow2~\cite{zimmermann2024virchow2} & 53.3 &  69.8 \\
Gigapath-SE~\cite{gigapath} & Gigapath~\cite{gigapath} &  48.3 &  66.9  \\
\midrule
\multirow{3}{*}{Tangle~\cite{jaume2024transcriptomics}} 
 & H-optimus-1 & 60.4 & 68.7   \\
&  UNI2-h & 53.8 &  69.5  \\
&  CONCHv1.5 & 60.6 &  71.6  \\

\midrule
\multirow{3}{*}{TICON$_{{tangle}}$}  
 & H-optimus-1 & 62.6 & 72.5 \\
&  UNI2-h & 58.9 &  70.3  \\
&  CONCHv1.5  & \textbf{63.8} & \textbf{72.7} \\

\bottomrule

\end{tabular}
}
\end{table}

\myparagraph{Evaluation}
To evaluate performance, we train a linear probe on the BRACS subtyping (2 tasks) and CPTAC mutation prediction (25 tasks) benchmarks from Patho-Bench~\cite{trident-pathobench}, following its protocol. As shown in Table~\ref{tab:slide-foundation-model}, we compare TICON$_{{tangle}}$ against two baselines: (i) Tangle trained directly on the original non-contextual tile embeddings and (ii) other state-of-the-art (SOTA) slide encoders.

\myparagraph{Results}
We observe that TICON$_{{tangle}}$ significantly outperforms the standard Tangle baseline for all tile encoders, with improvements of up to 5.1\% on BRACS and 3.8\% on CPTAC. This result highlights the critical importance of tile contextualization prior to slide-level aggregator pretraining. Furthermore, our TICON$_{{tangle}}$ model, pretrained using only 11K TCGA WSIs, outperforms all other slide encoders, even those trained on magnitudes more data. For comparison, TITAN~\cite{TITAN} uses 350K WSIs, Gigapath-SE~\cite{gigapath} uses 171K WSIs, and PRISM~\cite{shaikovski2024prism} uses 587K WSIs.

\subsection{Adapting TICON to unseen tile encoders}
\label{sec:results-unseen-tile-encoders}

We evaluate TICON's ability to adapt to unseen tile encoders, specifically Virchow2~\cite{zimmermann2024virchow2} and Gigapath~\cite{gigapath}. We introduce TICON$_{\text{tune}}$, which represents the TICON model adapted using our proposed method (Sec.~\ref{sec:ticon_adaptation}). This adaptation freezes the shared Encoder ($\mathcal{E}$) and Cross-Decoder ($\mathcal{D}$), training only new, lightweight input ($\rho_u$) and output ($\psi_u$) projectors for Virchow2 and Gigapath, respectively. In Table~\ref{tab:unseen-tile-encoders}, we compare TICON$_{tune}$ against original non-contextual tile embeddings.

\begin{table}[!htbp]
\centering
{
\caption{Adapting to unseen tile encoders. $\text{TICON}_{tune}$ represents training only projection layers to incorporate unseen tile encoders. Individual aggregators are pretrained with TANGLE prior to linear probing for slide-level tasks.}
\label{tab:unseen-tile-encoders}
\resizebox{0.9\columnwidth}{!}{%
\begin{tabular}{cccccc}
\toprule

  &  \textbf{Method}  & \multicolumn{2}{c}{\textbf{Tile-level tasks}}   & \multicolumn{2}{c}{\textbf{Slide-level tasks}} \\

  &   & \textbf{HEST} & \textbf{CATCH}   & \textbf{BRACS} & \textbf{CPTAC}  \\

   \midrule

  \multirow{2}{*}{\rotatebox[origin=c]{0}{\small \textbf{Virchow2}}} & Tile-encoder only &  0.398 & 86.0 & 57.6 & 71.4  \\
  & $\text{TICON}_{tune}$ & \textbf{0.411} & \textbf{86.8} & \textbf{64.2} & \textbf{71.9} \\

   \midrule

  \multirow{2}{*}{\rotatebox[origin=c]{0}{\small \textbf{Gigapath}}} &  Tile-encoder only &  0.385 & 83.0 & 58.4 & 68.9 \\
  & $\text{TICON}_{tune}$ & \textbf{0.396} & \textbf{83.2} & \textbf{59.9} & \textbf{70.3} \\

\bottomrule
\end{tabular}}}

\end{table}

We observe that TICON$_{tune}$ consistently outperforms the non-contextual baseline for both Virchow2 and Gigapath. TICON$_{tune}$ efficiently adapts the \textit{single, existing} core model ($\mathcal{E}$ and $\mathcal{D}$) by learning only new, lightweight projectors ($\rho_u$ and $\psi_u$). This approach is highly parameter-efficient and retains all capabilities for the original encoders used during pretraining.

\subsection{Ablations}
\label{sec:ablations}

Here, we provide ablations on the effect of our omni-pretraining vs. pretraining tile contextualizer independently for each tile encoder. We then study the choice of aggregator for slide-level tasks.

\myparagraph{Individual vs. Omni Pretraining}
In Table~\ref{tab:ablations-omni}, we compare the performance of our unified TICON, pretrained with the proposed \textit{omni feature masked modeling} strategy, against TICON$_{ind}$. TICON$_{ind}$ represents a baseline where a separate model was trained individually for each tile encoder using standard masked modeling (i.e., a single-target, self-reconstruction objective). We observe that the omni-trained TICON outperforms the individually-trained TICON$_{ind}$ models in 9 out of 12 comparisons across different encoders and tasks. This result highlights that our omni-training strategy not only provides the flexibility of a single universal contextualizer but also boosts performance.
Furthermore, TICON$_{ind}$ also consistently outperforms the non-contextual embeddings (``Tile-encoder only''), confirming that context-based pretraining is beneficial even without the omni-training strategy.

\begin{table}[!htbp]
\centering
{
\caption{Comparison of individual pretrained ($\text{TICON}_{ind}$) and omni pretrained model (TICON) on tile-level and slide-level tasks. $\text{TICON}_{ind}$ refers to training TICON from scratch on embeddings of the individual tile-encoder. Individual aggregators are pretrained with TANGLE prior to linear probing for slide-level tasks.}
\resizebox{0.9\columnwidth}{!}{%
\label{tab:ablations-omni}
\begin{tabular}{cccccc}
\toprule

 & &  \textbf{METHOD} & \textbf{H-optimus-1} & \textbf{UNI2-h} & \textbf{CONCHv1.5} \\
  
   \midrule

\multirow{6}{*}{\rotatebox[origin=c]{90}{\textbf{Tile-level tasks}}}  & \multirow{3}{*}{\rotatebox[origin=c]{90}{\small \textbf{HEST}}} & Tile-encoder only   & 0.422 & 0.414 & 0.379 \\

  \addlinespace[1pt]
  \cline{3-6}
\addlinespace[2pt]
   
& &   w/ $\text{TICON}_{ind}$   &  0.422 & 0.411 & \textbf{0.381} \\

& &  w/ TICON  & \textbf{0.427} & \textbf{0.419} &  \textbf{0.381 }\\

  \addlinespace[1pt]
  \cline{2-6}
\addlinespace[2pt]

& \multirow{3}{*}{\rotatebox[origin=c]{90}{\small \textbf{CATCH}}} & Tile-encoder only   & 86.2 & 85.5 & 81.6 \\

  \addlinespace[1pt]
  \cline{3-6}
\addlinespace[2pt]

&  &  w/ $\text{TICON}_{ind}$     & 86.8 & 86.3 & 83.4 \\

& &   w/ TICON    &\textbf{ 87.6} & \textbf{86.9} & \textbf{84.9} \\

   \midrule

\multirow{6}{*}{\rotatebox[origin=c]{90}{\textbf{Slide-level tasks}}}  & \multirow{3}{*}{\rotatebox[origin=c]{90}{\small \textbf{BRACS}}} & Tile-encoder only   & 60.4 & 53.8 & 60.6 \\

  \addlinespace[1pt]
  \cline{3-6}
\addlinespace[2pt]
   
  & &  w/ $\text{TICON}_{ind}$  & 61.9 & 58.7 & 61.9 \\

 &  &   w/ TICON     & \textbf{62.6} & \textbf{58.9} & \textbf{63.8} \\

     \addlinespace[1pt]
  \cline{2-6}
\addlinespace[2pt]

& \multirow{3}{*}{\rotatebox[origin=c]{90}{\small \textbf{CPTAC}}} & Tile-encoder only   & 68.7 & 69.5 & 71.6 \\

  \addlinespace[1pt]
  \cline{3-6}
\addlinespace[2pt]
   
 & &  w/ $\text{TICON}_{ind}$    & \textbf{72.5} & 69.9 & \textbf{73.9} \\
 &  &  w/ TICON     & \textbf{72.5} & \textbf{70.3} & 72.7 \\
      
\bottomrule
\end{tabular}}}
\end{table}

\myparagraph{Tangle vs Meanpool}
We evaluated two different aggregators for slide-level tasks with TICON: Meanpool and Tangle~\cite{jaume2024transcriptomics}, an attention-based aggregator. Both aggregators are used to extract a single global representation embedding for the whole WSI from contextualized tile embeddings. As observed in Table~\ref{tab:meanpool}, TICON$_{{tangle}}$ substantially outperforms TICON$_{{meanpool}}$ across all tile encoders. This highlights the effectiveness of using a Tangle-based aggregator to derive maximum benefit from the contextualized embeddings. This finding is consistent with recent evaluations of MAE pretrained encoders~\cite{przewikezlikowski2025beyond} in natural imaging, which confirm that attention-based aggregators more effectively represent an image at the global level compared to mean-pooling of patch tokens.

\begin{table}[!htbp]

\centering
\caption{Comparison of using Meanpool and Tangle aggregators to build a slide-level foundation model using TICON. We chose the Tangle aggregator because it clearly outperforms Meanpool in both datasets.}
\label{tab:meanpool}
\resizebox{0.9\columnwidth}{!}{
\begin{tabular}{@{}lc|cc@{}}
\toprule
\multirow{1}{*}{\textbf{Model}} &  \multirow{1}{*}{\textbf{Tile Encoder}} & 
\multicolumn{1}{c}{\textbf{BRACS}} &
\multicolumn{1}{c}{\textbf{CPTAC}} \\

& & 2 tasks & 25 tasks \\

\midrule
  \multirow{3}{*}{TICON$_{{meanpool}}$}
& Hoptimus-1 & 50.5 & 69.8 \\
& UNI-2 & 50.4 &  68.3 \\
& Conch-v15 & 51.8  &  69.1 \\

\midrule
\multirow{3}{*}{TICON$_{{tangle}}$}  
 & H-optimus-1 & 62.6 & 72.5 \\
&  UNI2-h & 58.9 &  70.3  \\
&  CONCHv1.5  & \textbf{63.8} & \textbf{72.7} \\

\bottomrule
\end{tabular}
}
\vspace{-5pt}
\end{table}

%% file: sec/conclusion.tex
\section{Conclusion}
We presented TICON, a universal tile contextualizer capable of processing embeddings from \textit{any} tile encoder. We demonstrated that its contextualized embeddings significantly improve performance across a diverse range of applications in computational pathology, establishing new state-of-the-art results on tile-level (HEST-Bench, THUNDER, CATCH) and slide-level (Patho-Bench) benchmarks. Our work highlights the critical importance of tile contextualization and demonstrates the benefits of a single, unified contextualizer that can leverage the complementary strengths of multiple tile encoders, each of which may excel at different downstream tasks. We believe that contextualized embeddings, as they provide a more coherent slide-level representation, have potential for other future applications. One example is generative modeling for pathology, where generating large pathology images requires slide-level context.

\section*{Acknowledgments}

This research used resources
of the Argonne Leadership Computing Facility, a U.S. Department of Energy (DOE) Office of
Science user facility at Argonne National Laboratory and is based on research supported by the
U.S. DOE Office of Science-Advanced Scientific Computing Research Program, under Contract No.
DE-AC02-06CH11357. This research was partially supported by National Institutes of Health (NIH) and National Cancer Institute (NCI) grants 1R21CA25849301A1, 1R01CA297843-01, 3R21CA258493-02S1, 1R03DE033489-01A1, National Science Foundation (NSF) grant 2442053,  NCI awards
1R21CA258493-01A1, 5U24CA215109, UH3CA225021,
U24CA180924, NSF grants IIS-2123920, IIS-2212046, IIS-2245652, the French National Agency for Research (Grants ANR-21-CE45-0007, ANR-23-CE45-0029) as well as the Health Data Hub (HDH) as part of the second edition of the France-Qu{\'e}bec call for projects Intelligence Artificielle en sant{\'e}, Stony Brook Profund 2022 seed funding, and generous support from Bob Beals and Betsy Barton. Parts of the computations have been performed using computational resources from GENCI-IDRIS (Grants 2025-
AD011015593, AD011016068). This research work was also supported by France 2030 funding managed by the National Research Agency (ANR) as part of IA CLUSTER program, reference ANR-23-IACL-0003 – DATAIA CLUSTER. The content is solely the responsibility of the authors and does not necessarily represent the official views of the National Institutes of Health.

%% file: sec/X_suppl.tex
\clearpage
\maketitlesupplementary
\appendix

This appendix presents the following materials: 
\begin{itemize}

    \item Additional  experiments and results 
    (Sec. \ref{additional_ablations})

    \begin{itemize}
    \item Effect of no multi-target pretraining
    \item Benchmarking against existing Slide Encoders
    \item Ablation on TICON’s Architecture
    \item Extended Analysis on HEST-Bench
    \item TransMIL as Aggregator in TANGLE Pretraining

\end{itemize}

    \item Additional implementation details
    (Sec. \ref{implementation_details_additional})

    \begin{itemize}
    \item TANGLE Setup
    \item Harmonizing the Field of View for Tile Encoders
    \item CATCH Data Curation
    \item Evaluation setting
    \item TICON's architecture and pretraining setting
    \end{itemize}

    \item Additional visualizations - 
    \begin{itemize}
    \item Overview of TICON’s multi-target versus single-target pretraining paradigms in Fig.~\ref{fig:cross_target_vs_no_cross_target_TICON}.
    \item Overview of TICON’s inference modes in Fig.~\ref{fig:isolated_vs_contextualized}. 
    \end{itemize}

\end{itemize}

\section{Additional experiments and results}
\label{additional_ablations}

\begin{enumerate}

\item \textbf{{Effect of no multi-target pretraining:}} We validate the effectiveness of our proposed Omni-Feature Masked Modeling (OFMM) objective in Table~\ref{tab:no_cross_target_unified_training}. Specifically, we compare our default TICON, trained to reconstruct embeddings from multiple diverse tile encoders (Multi-target), against a variant trained solely to reconstruct the input encoder's own features (Single-target). Fig.~\ref{fig:cross_target_vs_no_cross_target_TICON} describes the differences in multi-target and single-target pretraining. Both models are trained within the same unified pretraining framework. We observe that multi-target training generally outperforms the single-target baseline, surpassing it in 7 out of 12 comparisons across HEST-Bench and CATCH tasks. More importantly, TICON with multi-target prediction consistently improves upon the non-contextual (tile-encoder only) baseline in all 12 instances (12/12), whereas the single-target variant fails to do so in 2 cases (10/12). Furthermore, the overall state-of-the-art performance for each task is consistently achieved by TICON with the default multi-target objective. This confirms that compelling the model to predict varied semantic ``views'' of the same tile encourages the learning of a more robust and generalized contextual representation.

\noindent \textit{\textbf{Takeaway:} Our Omni-Feature Masked Modeling (OFMM) benefits from the Multi-target prediction objective, yielding more robust representations than Single-target prediction.}

\begin{table}[!htbp]
\centering
{
\caption{Comparison with no multi-target pretraining in TICON evaluated on tile-level tasks with slide context. PCC reported for HEST and F1-score for CATCH. Individual aggregator is pretrained with TANGLE prior to linear probing for slide-level tasks. \boxed{\cdot} denotes the state-of-the-art on the respective tasks.}
\label{tab:no_cross_target_unified_training}
\resizebox{1\columnwidth}{!}{%
\begin{tabular}{cccccc}
\toprule

&  & \textbf{Multi-target} & \textbf{H-optimus-1} & \textbf{UNI2-h} & \textbf{CONCHv1.5} \\

   \midrule

\multirow{6}{*}{\rotatebox[origin=c]{90}{\textbf{Tile-level tasks}}}  & \multirow{3}{*}{\rotatebox[origin=c]{90}{HEST}} & \multicolumn{1}{c}{Tile-encoder only }   & 0.422 & 0.414 & 0.379 \\

  \addlinespace[1pt]
  \cline{3-6}
\addlinespace[2pt]

 & & \ding{55} & 0.422 & 0.415 & \textbf{0.387} \\
 &  & \ding{51} & \boxed{\textbf{0.427}} & \textbf{0.419} & 0.381 \\

  \addlinespace[1pt]
  \cline{2-6}
\addlinespace[2pt]

&\multirow{3}{*}{\rotatebox[origin=c]{90}{CATCH}} & \multicolumn{1}{c}{Tile-encoder only }   & 86.2 & 85.5 & 81.6 \\

  \addlinespace[1pt]
  \cline{3-6}
\addlinespace[2pt]

 &  & \ding{55} & \boxed{\textbf{87.6}} & \textbf{87.1} & 84.0 \\
 &   & \ding{51}  & \boxed{\textbf{ 87.6}} & 86.9 & \textbf{84.9} \\

   \midrule

\multirow{6}{*}{\rotatebox[origin=c]{90}{\textbf{Slide-level tasks}}}  & \multirow{3}{*}{\rotatebox[origin=c]{90}{BRACS}} & \multicolumn{1}{c}{Tile-encoder only }   & 60.4 & 53.8 & 60.6 \\

  \addlinespace[1pt]
  \cline{3-6}
\addlinespace[2pt]

 &  & \ding{55}    & 58.4 & \textbf{60.0} & 62.4 \\
& & \ding{51}    & \textbf{62.6} & 58.9 & \boxed{\textbf{63.8}} \\

  \addlinespace[1pt]
  \cline{2-6}
\addlinespace[2pt]

& \multirow{3}{*}{\rotatebox[origin=c]{90}{CPTAC}} & \multicolumn{1}{c}{Tile-encoder only }   & 68.7 & 69.5 & 71.6 \\

  \addlinespace[1pt]
  \cline{3-6}
\addlinespace[2pt]

 &   & \ding{55}    & 72.1 & \textbf{70.7} & 72.4 \\
&   & \ding{51}    & \textbf{72.5} & 70.3 & \boxed{\textbf{72.7}} \\
      
\bottomrule
\end{tabular}}}
\end{table}

\item \textbf{Benchmarking against existing Slide Encoders:} In Table~\ref{tab:extension_of_slide_encoders}, we evaluate the contextualization capabilities of prior slide-level foundation models, specifically TITAN~\cite{TITAN} and Gigapath-SE~\cite{gigapath}, benchmarking them against TICON. For tile-level evaluation, we pass the WSI's tile embeddings (using CONCHv1.5 for TITAN and Gigapath for Gigapath-SE) through their respective slide encoders.

\begin{table}[!hbp]
\centering
{
\caption{Comparison of our TICON with TITAN and Gigapath-SE (slide encoder) on tile-level and slide-level tasks. $^\dagger$ For Gigapath tile encoder, we report our results with $\text{TICON}_{tune}$. Individual aggregator is pretrained with TANGLE prior to linear probing for slide-level tasks. \boxed{\cdot} denotes the state-of-the-art on the respective tasks.}

\resizebox{1\columnwidth}{!}{%
\label{tab:extension_of_slide_encoders}
\begin{tabular}{ccccccc}
\toprule

 & &  \textbf{METHOD} & \textbf{H-optimus-1} & \textbf{UNI2-h} & \textbf{CONCHv1.5} & \textbf{Gigapath} \\
  
   \midrule
   \midrule

\multirow{4}{*}{\rotatebox[origin=c]{90}{\textbf{Tile tasks}}}  & \multirow{4}{*}{\rotatebox[origin=c]{90}{\small CATCH}} & Tile-encoder only   & 86.2 & 85.5 & 81.6 & 83.0 \\

  \addlinespace[1pt]
  \cline{3-7}
\addlinespace[2pt]

& &   w/ TITAN   &  \small NA & \small NA & 84.4 & \small NA  \\

    & &   w/ Gigapath-SE   &  \small NA & \small NA & \small NA & 82.9  \\

& &   w/ TICON    & \boxed{\textbf{ 87.6}} & \textbf{86.9} & \textbf{84.9} & \textbf{83.2}$^\dagger$ \\

   \midrule
   \midrule

\multirow{8}{*}{\rotatebox[origin=c]{90}{\textbf{Slide-level tasks}}}  & \multirow{4}{*}{\rotatebox[origin=c]{90}{\small {BRACS}}} & Tile-encoder only   & 60.4 & 53.8 & 60.6 & 58.4  \\

  \addlinespace[1pt]
  \cline{3-7}
\addlinespace[2pt]
   
& &   w/ $\text{TITAN}_{tangle}$   &  \small NA & \small NA & 63.7 & \small NA  \\
 & &   w/ $\text{Gigapath-SE}_{tangle}$   &  \small NA & \small NA & \small NA & 59.1  \\

 &  &   w/ $\text{TICON}_{tangle}$     & \textbf{62.6} & \textbf{58.9}  & \boxed{\textbf{63.8}} & \textbf{59.9}$^\dagger$ \\

     \addlinespace[1pt]
  \cline{2-7}
\addlinespace[2pt]
  \cline{2-7}
& \multirow{4}{*}{\rotatebox[origin=c]{90}{\small {CPTAC}}} & Tile-encoder only   & 68.7 & 69.5 & 71.6 & 68.9  \\

  \addlinespace[1pt]
  \cline{3-7}
\addlinespace[2pt]
   
 & &   w/ $\text{TITAN}_{tangle}$   &  \small NA & \small NA & 71.4 & \small NA  \\
 & &   w/ $\text{Gigapath-SE}_{tangle}$   &  \small NA & \small NA & \small NA & 69.6  \\

 &  &  w/ $\text{TICON}_{tangle}$     & \textbf{72.5} & \textbf{70.3} & \boxed{\textbf{72.7}} & \textbf{70.3}$^\dagger$  \\
      \bottomrule

\bottomrule
\end{tabular}}}
\end{table}

\begin{table*}[!htbp]
    \centering
    \caption{\textbf{Ablation Studies.} We evaluate tile-level tasks with slide context, reporting the Pearson correlation coefficient (PCC) for HEST-Bench and the F1-score for CATCH. We compare the baseline tile encoders against our TICON framework across three variations: (a) Masking ratio ($m_r$), (b) Slide Encoder's embedding size ($D$), and (c) Slide Encoder's depth ($l$). $^\dagger$ represents the default parameter.}
    \label{tab:all_ablations}
    
    \begin{subtable}[t]{0.32\textwidth}
        \centering
        \caption{Masking ratio ($m_r$)}
        \label{tab:ablation_masking}
        \resizebox{\linewidth}{!}{%
        \begin{tabular}{l c c c}
            \toprule
            \textbf{Method} & $m_r$ & \textbf{HEST} & \textbf{CATCH} \\
            \midrule
            H-optimus-1 & NA & 0.422 & 86.2 \\
            \addlinespace[1pt] \cline{2-4} \addlinespace[2pt]
            \multirow{2}{*}{w/ TICON} & 90\% & \textbf{0.428} & 87.5 \\
             & 75\%$^\dagger$ & 0.427 & \textbf{87.6} \\
            \midrule
            UNI2-h & NA & 0.414 & 85.5 \\
            \addlinespace[1pt] \cline{2-4} \addlinespace[2pt]
            \multirow{2}{*}{w/ TICON} & 90\% & \textbf{0.420} & 86.3 \\
             & 75\%$^\dagger$ & 0.419 & \textbf{86.9} \\
            \midrule
            CONCHv1.5 & NA & 0.379 & 81.6 \\
            \addlinespace[1pt] \cline{2-4} \addlinespace[2pt]
            \multirow{2}{*}{w/ TICON} & 90\% & 0.371 & 84.8 \\
             & 75\%$^\dagger$ & \textbf{0.381} & \textbf{84.9} \\
            \bottomrule
        \end{tabular}}
    \end{subtable}
    \hfill
    \begin{subtable}[t]{0.32\textwidth}
        \centering
        \caption{Embedding size ($D$)}
        \label{tab:ablation_dim}
        \resizebox{\linewidth}{!}{%
        \begin{tabular}{l c c c}
            \toprule
            \textbf{Method} & $D$ & \textbf{HEST} & \textbf{CATCH} \\
            \midrule
            H-optimus-1 & NA & 0.422 & 86.2 \\
            \addlinespace[1pt] \cline{2-4} \addlinespace[2pt]
            \multirow{3}{*}{w/ TICON} & 384 & 0.425 & 87.1 \\
             & 768 & 0.423 & \textbf{87.9} \\
             & 1536$^\dagger$ & \textbf{0.427} & 87.6 \\
            \midrule
            UNI2-h & NA & 0.414 & 85.5 \\
            \addlinespace[1pt] \cline{2-4} \addlinespace[2pt]
            \multirow{3}{*}{w/ TICON} & 384 & 0.417 & \textbf{87.1} \\
             & 768 & 0.417 & 86.7 \\
             & 1536$^\dagger$ & \textbf{0.419} & 86.9 \\
            \midrule
            CONCHv1.5 & NA & 0.379 & 81.6 \\
            \addlinespace[1pt] \cline{2-4} \addlinespace[2pt]
            \multirow{3}{*}{w/ TICON} & 384 & 0.362 & 84.6 \\
             & 768 & 0.378 & \textbf{85.1} \\
             & 1536$^\dagger$ & \textbf{0.381} & 84.9 \\
            \bottomrule
        \end{tabular}}
    \end{subtable}
    \hfill
    \begin{subtable}[t]{0.302\textwidth}
        \centering
        \caption{Depth ($l$)}
        \label{tab:ablation_depth}
        \resizebox{\linewidth}{!}{%
        \begin{tabular}{l c c c}
            \toprule
            \textbf{Method} & $l$ & \textbf{HEST} & \textbf{CATCH} \\
            \midrule
            H-optimus-1 & NA & 0.422 & 86.2 \\
            \addlinespace[1pt] \cline{2-4} \addlinespace[2pt]
            \multirow{3}{*}{w/ TICON} & 4 & 0.425 & \textbf{87.8} \\
             & 6$^\dagger$ & \textbf{0.427} & 87.6 \\
             & 12 & \textbf{0.427} & 86.4 \\
            \midrule
            UNI2-h & NA & 0.414 & 85.5 \\
            \addlinespace[1pt] \cline{2-4} \addlinespace[2pt]
            \multirow{3}{*}{w/ TICON} & 4 & 0.418 & 86.4 \\
             & 6$^\dagger$ & \textbf{0.419} & 86.9 \\
             & 12 & 0.418 & \textbf{87.3} \\
            \midrule
            CONCHv1.5 & NA & 0.379 & 81.6 \\
            \addlinespace[1pt] \cline{2-4} \addlinespace[2pt]
            \multirow{3}{*}{w/ TICON} & 4 & \textbf{0.383} & \textbf{84.9} \\
             & 6$^\dagger$ & 0.381 & \textbf{84.9} \\
             & 12 & 0.376 & 84.4 \\
            \bottomrule
        \end{tabular}}
    \end{subtable}
\end{table*}

\noindent We discard the global `[CLS]' token and utilize the output tile embeddings for the CATCH dataset. For slide-level tasks, we follow our established protocol by training a TANGLE aggregator on the contextualized tile embeddings produced by each slide encoder. We make two key observations: (1) Both TITAN and Gigapath-SE, when acting as contextualizers, yield improvements on 2 out of 3 tasks compared to their non-contextual baselines. (2) TICON consistently outperforms both slide encoders, even when restricted to the same input tile encoder. Furthermore, TICON achieves superior overall performance on tile-level tasks by leveraging its flexibility to process state-of-the-art tile encoders (e.g., H-optimus-1 on CATCH), a capability lacking in the baseline slide encoders, which are tied to a specific input tile encoder.

\noindent We further highlight the efficacy of our omni-feature multi-target masked pretraining by comparing the results in Table~\ref{tab:no_cross_target_unified_training} and Table~\ref{tab:extension_of_slide_encoders}. We observe that TICON outperforms TITAN across all 3 tasks (using the same CONCHv1.5 input) \textit{only} when the multi-target objective is employed in pretraining. In contrast, without multi-target pretraining, TICON falls behind TITAN on 2 of the 3 tasks, underscoring the critical importance of cross-encoder alignment. Consequently, our multi-target approach enables highly data-efficient pretraining: despite relying on only 11K WSIs from the open-source TCGA dataset, TICON surpasses baselines like TITAN, which benefit from pretraining on over 350K WSIs.

\noindent \textit{\textbf{Takeaway:} While current slide encoders offer gains when used as contextualizers, TICON's multi-target pretraining delivers superior performance and data efficiency, consistently outperforming baselines even with magnitudes less pretraining data.}

\item \textbf{Ablation on TICON's Architecture:} We conduct a comprehensive ablation study on TICON's architectural design choices in Table~\ref{tab:all_ablations}:
\begin{itemize}
    \item \textbf{Masking ratio ($m_r$):} We compare masking ratios of 90\% and 75\%. While the model remains robust even at a high masking ratio of 90\%, we observe that $m_r = 75\%$ generally yields the most stable and optimal performance across tasks; thus, we adopt it as our default.
    \item \textbf{Embedding size ($D$):} We evaluate TICON's shared embedding space with dimensions of $D \in \{384, 768, 1536\}$. Increasing the capacity to $D=1536$ results in consistent performance gains across all encoders on HEST-Bench. On CATCH, while smaller dimensions lead to inconsistent rankings when inferred with different tile encoders, $D=1536$ provides the most robust and stable performance, justifying its selection for our final model.
    \item \textbf{Depth ($l$):} We vary the depth of the shared encoder ($l \in \{4, 6, 12\}$). We find that a depth of 6 layers ($l=6$) offers the optimal trade-off between performance and computational efficiency. Similar to our observations regarding embedding size, depths of 4 or 12 result in fluctuating performance rankings on HEST-Bench and CATCH, whereas $l=6$ consistently provides stable and high performance across both tasks and diverse tile encoders. 
    \\
    
\end{itemize}

\item \textbf{Extended Analysis on HEST-Bench:} Table~\ref{tab:hest_with_meanpool} presents an extended analysis of the gene expression prediction task on HEST-Bench. In the main paper, our baselines relied solely on the `[CLS]' token from the tile encoder and so do TICON's input. Here, we adopt the more rigorous protocol common in recent literature, where the `[CLS]' token is concatenated with the mean-pooled patch tokens from the tile encoder. We then concatenate TICON's contextualized output (which corresponds to the `[CLS]' tokens input) with the non-contextual `[CLS]' and mean-pooled patch tokens.

\begin{table}[!htbp]
\centering
{
\caption{HEST performance with CLS only and with CLS+meanpool concatenated. }
\label{tab:hest_with_meanpool}
\resizebox{0.8\columnwidth}{!}{%
\begin{tabular}{l c }
\toprule
    \textbf{Method} & \textbf{HEST-Bench} (9 tasks) \\
    \midrule
    H-optimus-1   (CLS)  & 0.422 \\
   w/  TICON & 0.427 \\

    H-optimus-1 (CLS+meanpool)     & 0.431 \\
   w/  TICON & \textbf{0.437} \\

    \midrule
    
    UNI2-h    (CLS)   & 0.414  \\
 w/    TICON  & 0.419 \\
        UNI2-h (CLS+meanpool)      & 0.431 \\
 w/    TICON  & \textbf{0.437} \\
 
    \midrule
    CONCHv1.5  & 0.379 \\
   w/  TICON  & \textbf{0.381}  \\
   
\bottomrule
\end{tabular}}}
\end{table}

We observe that while including mean-pooled features significantly boosts the baseline performance, particularly for UNI2-h, TICON successfully maintains its lead, pushing the state-of-the-art frontier across all tile encoders. It is worth noting that our current TICON model is pretrained using only the `[CLS]' token; explicitly incorporating mean-pooled embeddings during TICON's pretraining could potentially yield further performance improvements on benchmarks like HEST and other tasks.

\noindent \textit{\textbf{Takeaway:} TICON's slide-level contextualization provides complementary information to local feature aggregation, consistently boosting performance even against strengthened baselines that incorporate mean-pooled representations.}  \\

\item \textbf{TransMIL as Aggregator in TANGLE Pretraining:} We investigate the choice of the underlying aggregation mechanism for TANGLE slide-level contrastive pretraining in Table~\ref{tab:transmil}. We compare our default Attention-Based MIL (ABMIL) aggregator against the Transformer-based TransMIL~\cite{shao2021transmil}. Evaluating on Patho-Bench (BRACS and CPTAC), we observe that the simpler ABMIL aggregator consistently outperforms TransMIL. This indicates that learning contextualization via a Transformer within a multimodal contrastive setting may be suboptimal. Crucially, this highlights the distinct advantage of TICON as a standalone contextualizer: it demonstrates that learning context through a dedicated masked modeling objective (reconstructing visual embeddings) is more effective than attempting to derive context implicitly through alignment with auxiliary modalities like gene expression. Consequently, pretraining an aggregator on these TICON-contextualized embeddings leads to a state-of-the-art slide-level foundation model, despite using only 11K WSIs. This finding suggests that large-scale whole slide-encoder pretraining methods like TITAN and PRISM, which access hundreds of thousands of WSI-report pairs, could potentially achieve even greater performance by integrating TICON's contextualized embeddings into their vision-language frameworks.

\textit{\textbf{Takeaway:} Our Contextualizer as a standalone stage benefits Slide-level pretraining with TANGLE over trying to train the contextualizer (transformer based MIL) too with the multimodal contrastive objective.}

\begin{table}[!htbp]

\centering
\caption{Comparison with TransMIL as aggregator for non-contextual tile embeddings with Patho-bench~\cite{trident-pathobench}. Following Pathobench, we report balanced accuracy for BRACS and AUC for CPTAC. TICON$_{{tangle}}$ improves over Tangle across both aggregators. Note that a multi-head version of ABMIL is used as aggregator in this study for baselines and ours.}
\label{tab:transmil}
\resizebox{0.9\columnwidth}{!}{
\begin{tabular}{@{}lcc|cc@{}}
\toprule
\multirow{1}{*}{\textbf{Model}} &  \multirow{1}{*}{\textbf{Aggregator}} &  \multirow{1}{*}{\textbf{Tile Encoder}} & 
\multicolumn{1}{c}{\textbf{BRACS}} &
\multicolumn{1}{c}{\textbf{CPTAC}} \\

& & & 2 tasks & 25 tasks \\

\midrule
\multirow{2}{*}{Tangle~\cite{jaume2024transcriptomics}} 
& ABMIL &  CONCHv1.5 & 60.6 &  71.6  \\

& TransMIL&  CONCHv1.5 & 60.3 &  69.0  \\

\midrule
\multirow{1}{*}{TICON$_{{tangle}}$}  
& ABMIL&  CONCHv1.5  & \textbf{63.8} & \textbf{72.7} \\

\bottomrule

\end{tabular}
}
\end{table}

\item \textbf{Unified vs. Individual TANGLE Pretraining:} In our primary evaluation (main paper Table~\ref{tab:slide-foundation-model}), we pretrained a separate MIL aggregator using TANGLE objective for each tile encoder for baseline and for it's TICON's contextualized embeddings. In Table~\ref{tab:unified_vs_individual_tangle_pretraining}, we explore a \textit{unified} approach: training a single TANGLE model on the contextualized embeddings from all three pretraining tile encoders (UNI2-h, CONCHv1.5, and H-optimus-1). Since TICON maps all inputs to a shared embedding dimension $D$, we train this unified MIL aggregator by randomly sampling the source tile encoder for each WSI within a batch.

\begin{table}[!htbp]

\centering
\caption{Comparison of unified vs. individual tile encoder (default) TANGLE pretraining. Evaluation on slide-Level tasks with Patho-Bench~\cite{trident-pathobench}. Following Patho-Bench, we report balanced accuracy for BRACS and AUC for CPTAC.}

\label{tab:unified_vs_individual_tangle_pretraining}
\resizebox{0.8\columnwidth}{!}{
\begin{tabular}{@{}lc|cc@{}}
\toprule
\multirow{1}{*}{\textbf{Model}} &  \multirow{1}{*}{\textbf{Tile Encoder}} & 
\multicolumn{1}{c}{\textbf{BRACS}} &
\multicolumn{1}{c}{\textbf{CPTAC}} \\

& & 2 tasks & 25 tasks \\

\midrule 
\multirow{3}{*}{Tangle~\cite{jaume2024transcriptomics}} 
 & H-optimus-1 & 60.4 & 68.7   \\
&  UNI2-h & 53.8 &  69.5  \\
&  CONCHv1.5 & 60.6 &  71.6  \\

\midrule
\multirow{3}{*}{TICON$_{{tangle\_{unified}}}$}  
 & H-optimus-1 & 61.3 & 72.5 \\
&  UNI2-h & 58.8 &  70.0 \\
&  CONCHv1.5  & 60.6 & 71.9 \\

\midrule
\multirow{3}{*}{TICON$_{{tangle}}$}  
 & H-optimus-1 & 62.6 & 72.5 \\
&  UNI2-h & 58.9 &  70.3  \\
&  CONCHv1.5  & \textbf{63.8} & \textbf{72.7} \\

\bottomrule

\end{tabular}
}
\end{table}

We observe that while the individually pretrained aggregators generally outperform the unified model, the unified TANGLE still surpasses the non-contextual baselines in 5 out of 6 cases and matches performance in the remaining one. Future work could further optimize this unified MIL aggregator pretraining, for instance, through homogeneous batch sampling (fixing the tile encoder per batch) or by incorporating tile-encoder-specific projection MLP heads similar to TICON's design.

\noindent \textit{\textbf{Takeaway:} While unified training proved beneficial for TICON's masked modeling objective, we found that for contrastive objectives like TANGLE (WSI-gene alignment), individual aggregator training remains superior. The unified TANGLE training requires further exploration.}

\end{enumerate}

\section{Additional implementation details}
\label{implementation_details_additional}

\begin{enumerate}

\item{\textbf{TANGLE Setup:}} For slide-level tasks, we pretrain a multi-head (2 heads) ABMIL aggregator to pool the tile embeddings. We fix the hidden dimension at 512 and incorporate a 3-layer pre-attention MLP, following the architecture specified in the TANGLE pan-cancer pretraining codebase. For the gene modality, we employ a 3-layer MLP with a hidden dimension of 512. For all TANGLE pretraining runs, we use a batch size of 512 and fix the number of randomly sampled tokens per WSI to 4096 to enable batching~\cite{jaume2024transcriptomics}. Apart from these changes, we adopt all default hyperparameters from the original TANGLE implementation. \\

\item \textbf{Harmonizing the Field of View for Tile Encoders:} We observe that input tile resolutions vary significantly across encoders (e.g., $512 \times 512$ px for CONCHv1.5, $224 \times 224$ px for H-optimus-1 and Virchow2, and $256 \times 256$ px for UNI2-h and Prov-GigaPath). To harmonize these for our omni-pretraining, we standardize the base extraction area to $512 \times 512$ px. For native $512 \times 512$ encoders like CONCHv1.5, we extract embeddings directly. For encoders requiring smaller inputs, we subdivide the $512 \times 512$ area into four $256 \times 256$ quadrants, extract embeddings for each (resizing to $224 \times 224$ if necessary), and compute their mean. This ensures a one-to-one spatial correspondence across all tile encoders. Furthermore, this pooling strategy significantly enhances computational efficiency during inference for both TICON and the subsequent TANGLE aggregation by reducing the effective sequence length by a factor of four for all encoders except CONCHv1.5 (which natively operates at $512 \times 512$).

Crucially, since this mean-pooling strategy preserves the semantic distribution of the embedding space, TICON retains the flexibility at inference time to process either the aggregated representations or the original fine-grained embeddings directly. This capability is essential for benchmarks like HEST, which require gene expression predictions at the native resolution ($224 \times 224$ px at $20\times$). Similarly, for THUNDER, we process each tile at the native resolution required by different tile encoders and pass the resulting single embedding through TICON (where the sequence length is 1, effectively TICON acting as a deep MLP).

\textbf{Comparison with baseline Tile-encoder.} To ensure a fair comparison for the ``Tile Encoder Only'' baselines (excluding HEST and THUNDER), we apply the same aggregation methodology: non-$512 \times 512$ tile encoder outputs are mean-pooled prior to downstream usage (e.g., k-NN classification for CATCH or TANGLE pretraining). This ensures a strictly one-to-one comparison between the non-contextual tile embeddings and our TICON-contextualized counterparts.

\textbf{Comparison with baseline Slide-encoders.} In contrast, for baseline slide encoders such as Gigapath-SE~\cite{gigapath}, which expect fine-grained input embeddings (e.g., $256 \times 256$ px), we bypass TICON's pooling operation. As reported in Table~\ref{tab:extension_of_slide_encoders}, we train the TANGLE aggregator for Gigapath-SE by performing contextualization directly on these native resolution tiles. Conversely, since TITAN~\cite{TITAN} is designed to process CONCHv1.5 embeddings ($512 \times 512$ px), we apply it directly without modification. For the CATCH tile classification task, where ground-truth labels are defined at the $512 \times 512$ px resolution, we adapt the Gigapath-SE output by mean-pooling the contextualized embeddings of the corresponding $2 \times 2$ quadrant to yield a single representative embedding for each labeled $512 \times 512$ region. \\

\item \textbf{CATCH Data Curation:} We observe that despite the recent proliferation of histopathology benchmarks (e.g., HEST-Bench, THUNDER, and Patho-Bench), there remains a scarcity of datasets that enable tile-level evaluation within a full slide-level context. HEST-Bench serves as a notable exception: while originally proposed as a tile-level task, it retains the spatial coordinates for all tiles, allowing us to reformulate it as a tile-level task with full slide context available. This broader gap in the field primarily stems from the traditional ``bag-of-words'' paradigm, which typically treats tile-level feature extraction and slide-level aggregation as distinct, decoupled tasks. To bridge this gap and evaluate the impact of context on local predictions, we curate the CATCH dataset to support tile-level classification while retaining the spatial integrity of the Whole Slide Image. 

For this curation, we utilize the segmentation contours available in the CATCH~\cite{catch} dataset. From the original 13 classes, we exclude ``cartilage'' due to its low prevalence, retaining the remaining 12 classes. We compute the overlap of $512 \times 512$ tiles (at 20$\times$ magnification) with the original contours. A tile is assigned a label only if it is fully contained (100\% overlap) within a contour of one of the 12 classes. To ensure an unambiguous multi-class classification task, we discard any tiles that intersect with contours of multiple different classes. This process results in a dataset of 916,967 patches derived from 350 WSIs, split into 210 for training, 49 for validation, and 91 for testing. We evaluate performance using k-NN probing, selecting the optimal $k$ on the validation set and reporting the final performance on the test set. 

We plan to release this benchmark along with the whole curation pipeline. We anticipate that our research, by bridging the traditionally decoupled stages of tile embedding extraction and aggregation with a slide-encoder as contextualizer, will catalyze the creation of further benchmarks designed to evaluate the dense prediction capabilities of slide encoders, moving beyond solely global slide-level tasks.\\

\item \textbf{Evaluation Setting:} In this study, we strictly adhere to the default metrics and hyperparameters established by the respective benchmarks. 

For slide-level tasks, we utilize Patho-Bench, adopting the two subtyping tasks (coarse-grained and fine-grained) from BRACS and the 25 mutation prediction tasks from CPTAC. For both datasets, we employ linear probing with a balanced loss and a cost parameter of $0.5$, while keeping all other parameters at their defaults. We report performance using the specific metrics provided by the benchmark for each task.

For tile-level tasks, we follow specific established protocols:
\begin{itemize}
    \item \textbf{HEST-Bench:} We adopt the default setup of applying PCA to reduce dimensions to 256, followed by ridge regression.
    \item \textbf{THUNDER:} We utilize their default pipeline to report average k-NN results for the 12 original tasks and the 4 newly added SPIDER~\cite{nechaev2025spider} tasks.
    \item \textbf{CATCH:} Aligning with the THUNDER protocol, we employ the same k-NN based evaluation strategy.
\end{itemize}

Importantly, for these tile-level tasks, which do not require additional learnable parameters  (due to the use of PCA or non-parametric k-NN) with increase in feature size, we enhance the representation by concatenating the original non-contextual tile embeddings with their corresponding TICON-contextualized embeddings (or the output of TICON$_{iso}$ for THUNDER). Conversely, for slide-level tasks, to maintain consistency in TANGLE's input dimension, we utilize only the contextualized tile embeddings, discarding the original non-contextual inputs.

\item{\textbf{TICON's architecture and pretraining setting:}}

TICON was pretrained using the default hyperparameters listed in Table~\ref{tab:pretraining_hyperparameters} on a setup consisting of 8 $\times$ NVIDIA A100 40GB GPUs. Memory profiling during the training indicated a consumption of approximately 5GB per GPU. The entire pretraining was completed in 10 hours.

\begin{table}[!htbp]
\centering
{
\caption{Pretraining hyperparameters of TICON }
\label{tab:pretraining_hyperparameters}
\resizebox{0.7\columnwidth}{!}{%
\begin{tabular}{lc}
\toprule
    \textbf{Hyperparameter} & \textbf{Value} \\
    \midrule
    Batch size     & 1024 \\
    Optimizer & AdamW(0.9, 0.95) \\
    Learning rate & 2e-4 \\
    Weight decay & 0.05 \\
    Warmup iterations & 10K \\
    Total iterations & 100K \\
    Training dtype & bf16 \\
    Parallelism & FSDP \\
    Masking type & random \\
    Masking ratio ($m_r$) & 75\% \\
    Prediction ratio ($p_r$) & 25\% \\

\bottomrule
\end{tabular}}}
\end{table}

TICON pretraining consists of an Encoder (ViT 6 layers, 1536 embed dim) with 170 million (M) parameters and a Cross-Decoder (ViT 1 layer, 1536 embed dim) with 28M parameters. Additionally, the input and output projectors (2 layer MLP) each contribute up to 5M parameters. When adapting to unseen tile encoders, we only train the parameters of new projectors for 20K iterations.

\textbf{Partial prediction.} We only reconstruct a partial amount of the target embeddings during pretraining, based on a prediction ratio ($p_r$), rather than all of the masked embeddings. We choose this partial prediction strategy because our pretraining candidates contain a minimum of 55\% tiles with tissue, while the remaining tiles are invalid regions. We exclude these invalid regions from both the visible embeddings (input to the encoder) and the prediction targets (output of the decoder). Since we use a masking ratio ($m_r$) of $75\%$, we opt for a prediction ratio ($p_r$) of $25\%$ of the total embeddings. This restriction ensures that the total operated embeddings (visible 25\% in the encoder and target 25\% in the decoder) remains below the minimum tissue occupancy of $55\%$ in the pretraining candidates. A recent study $\cite{crossmae}$ demonstrated that a cross-attention-based decoder (Cross-Decoder) is better suited for partial prediction than its self-attention counterpart. Consequently, we adopted a Cross-Decoder for our pretraining architecture.

\begin{center}
 \begin{figure*}[!htbp]
 \centering
     \includegraphics[width=1\linewidth]{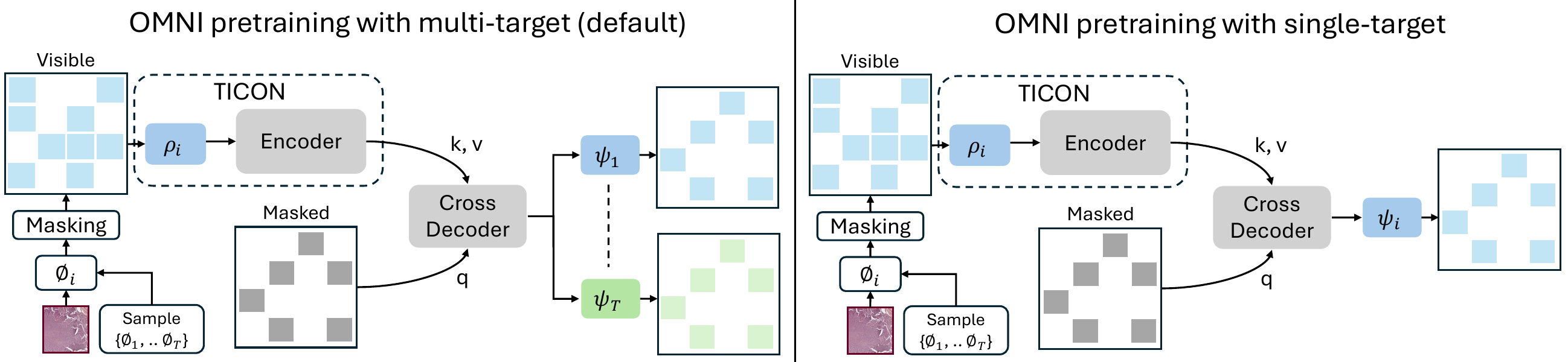} 
      \caption{Overview of TICON's multi-target versus single-target pretraining paradigms. \textbf{(Left) Default Multi-Target Pretraining:} At each iteration, we randomly sample a single input tile encoder. Its embeddings are projected and encoded, after which the decoder is tasked with reconstructing the masked embeddings for \textit{all} tile encoders (used in omni-pretraining) simultaneously. This mechanism enforces cross-encoder semantic alignment. \textbf{(Right) Single-Target Pretraining:} In this ablation setting, the model retains the capacity to process any encoder (omni-compatible) but lacks cross-target supervision. Specifically, the decoder is restricted to reconstructing only the masked embeddings of the input encoder itself. Thus, while the input encoder varies randomly across iterations, the target is always identical to the input, removing explicit cross-encoder prediction.}
     \label{fig:cross_target_vs_no_cross_target_TICON}
     \vspace{-6cm}
 \end{figure*}
 \end{center}

\begin{center}
 \begin{figure*}[!htbp]
 \centering
     \includegraphics[width=1\linewidth]{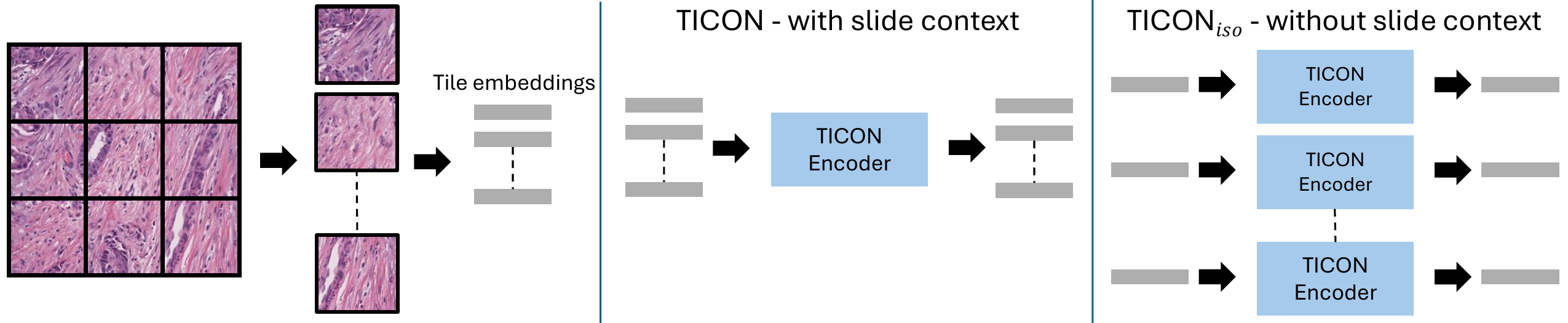} 
 \caption{Overview of TICON's inference modes. \textbf{(Left)} Standard preprocessing pipeline: tiling the WSI followed by embedding extraction. \textbf{(Middle) Contextualized Inference:} The default mode where the entire sequence of WSI tile embeddings is passed through the TICON Encoder. This allows the model to contextualize each tile with information from the full slide-level neighborhood. \textbf{(Right) Isolated Inference:} An alternative inference mode where a single tile embedding is passed through TICON independently. In this setting, the Transformer effectively functions as a deep MLP (sequence length of 1). Although not the primary design intent, we empirically discovered that TICON exhibits an emergent property in this mode, enhancing individual tile representations even when slide-level context is unavailable (e.g., in the THUNDER benchmark).}
     \label{fig:isolated_vs_contextualized}
 \end{figure*}
 \end{center}

\end{enumerate}